\documentclass{article}
\usepackage{spconf,amsmath,graphicx}

\usepackage{tabularx}
\usepackage{multirow}
\usepackage{verbatim}
\usepackage{xspace}
\usepackage{url}
\usepackage[hidelinks]{hyperref}
\usepackage{booktabs}
\def\F0{$F_0$\xspace}
\usepackage{siunitx}

\title{Controllable Speaking Styles Using a Large Language Model}

\name{Atli Sigurgeirsson\thanks{This work was supported in part by Huawei and the UKRI Centre for Doctoral Training in Natural Language Processing, funded by the UKRI (grant EP/S022481/1) and the University of Edinburgh, School of Informatics and School of Philosophy, Psychology \& Language Sciences. A short version of this work was presented as a non-peer-reviewed Late Breaking Report at the 12th ISCA Speech Synthesis Workshop.
}, Simon King}
\address{The Centre for Speech Technology Research, University of Edinburgh, UK}

\begin{document}
\ninept

\maketitle

\begin{abstract}
Reference-based Text-to-Speech (TTS) models can generate multiple, prosodically-different renditions of the same target text. Such models jointly learn a latent acoustic space during training, which can be sampled from during inference. Controlling these models during inference typically requires finding an appropriate reference utterance, which is non-trivial.

Large generative language models (LLMs) have shown excellent performance in various language-related tasks. Given only a natural language query text (the `prompt'), such models can be used to solve specific, context-dependent tasks. Recent work in TTS has attempted similar prompt-based control of novel speaking style generation. Those methods do not require a reference utterance and can, under ideal conditions, be controlled with only a prompt. But existing methods typically require a prompt-labelled speech corpus for jointly training a prompt-conditioned encoder. 

In contrast, we instead employ an LLM to directly suggest prosodic modifications for a controllable TTS model, using contextual information provided in the prompt. The prompt can be designed for a multitude of tasks. Here, we give two demonstrations: control of speaking style; prosody appropriate for a given dialogue context. The proposed method is rated most appropriate in 50\% of cases vs. 31\% for a baseline model.
\end{abstract}


\section{Introduction}
\label{sec:intro}
The same message can be spoken in many different ways \cite{wilson2006relevance}. The speaker chooses an appropriate prosodic rendition to encode a particular interpretation or to convey attitude and emotion \cite[p.\ 13]{cole2015prosody}. Some prosodic renditions are appropriate in a given context while others are not: choosing the right one can therefore be critical to deliver a particular meaning \cite{wagner2010experimental}. Typical end-to-end Text-To-Speech models only generate a `default prosody', which may be either perceived as inexpressive or as inappropriate to the context, which can affect comprehension \cite{govender2018using} and perceived naturalness. They have no means of \textit{selecting} the right prosodic rendition; they simply learn the prosodic distribution of the training corpus.

\section{Background}
\textit{Reference-based} TTS models can generate different prosodic renditions, given the same target text. These models learn a latent acoustic space during training that can then be sampled from during inference, to generate speech acoustically-similar to the chosen reference spoken utterance. Such models have been used for Prosody-Transfer (PT), proposed in \cite{pt_main} and subsequently used in \cite[e.g.]{battenberg2019effective,zaidi22b_interspeech, sigurgeirsson2022}, as well as for modelling speaking style \cite{wang2018style}. These models can generate expressive and context-appropriate speech, provided that the model is conditioned on an externally-provided and appropriately-chosen reference utterance. 

\textit{Large Language Models} \cite[e.g.]{brown2020language, ouyang2022training} are, when sufficiently large, capable few-shot learners in many language-related tasks \cite{brown2020language}. LLMs are very flexible because the target task can be fully defined using only natural language. The task description and the task itself (collectively known as the `prompt') are used to query the LLM for a solution without any fine-tuning or model parameter updates. LLMs have been used to solve tasks as diverse as text completion, answering factual questions, translation, grammar correction, etc \cite{brown2020language}.

Describing the target task using natural language offers an intuitive way of interacting with models which are otherwise entirely opaque. Prompt-controlled models have been employed for a wide spectrum of different tasks and modalities including \textit{prompt-to-image} \cite{ramesh2021zero} and \textit{prompt-to-music} \cite{agostinelli2023musiclm}. Recent work has attempted to apply this form of control to expressive speech generation \cite{guo2022prompttts, yang2023instructtts, Yoon2022LanguageME}. However, those methods require a \textit{prompt-labelled} speech corpus for jointly training a \textit{prompt-encoder} with the acoustic model. Very few such corpora exist.

TTS models such as FastSpeech-2 \cite{ren2019fastspeech}
and Daft-Exprt \cite{zaidi22b_interspeech} explicitly model acoustic correlates of prosody. These models offer interpretable   prosody modification: during inference, the predicted values of these acoustic correlates could, in principle, be tuned by a human expert to fulfill a prosodic requirement. However, this would be tedious and require a high level of expertise, making this approach infeasible for most applications.

In the current work, we explore whether an LLM can replace that human expert. We prompt \textit{InstructGPT} -- an LLM that has already been fine-tuned to follow natural language instructions \cite{ouyang2022training} -- to suggest context-appropriate modifications to the acoustic features in a modified FastSpeech-2 model. The proposed method does not require training of a prompt-encoder, or a style-labelled corpus. The flexibility of our design allows for changing the task of the LLM arbitrarily. We explore its capabilities when 1) prompted with a target speaking style described using natural language; and 2) when provided with the previous line in an expressive dialogue. 

To the best of our knowledge, the proposed method is novel and the first to prompt an LLM \textit{using only natural language} to solve a specific task in TTS.

\section{Related Work}
\label{sec:related}
Specifying target prosody by providing a reference utterance may be more intuitive for certain applications than requiring a detailed (e.g., per phone) specification of acoustic parameters. Reference-based TTS models, such as PT models, offer this type of control; however, choosing the right reference is problematic. A reference utterance from a different speaker often leads to \textit{source-speaker leakage} \cite{sigurgeirsson2022}. \textit{Feature entanglement} \cite{pt_main, battenberg2019effective} is a concern for cross-text prosody transfer. Both greatly complicate choosing an appropriate reference. Style token models \cite{wang2018style} learn a finite set of style \textit{tokens} by jointly training a reference encoder. The learned tokens capture perceptually-distinct speaking styles \cite{wang2018style} but require expressive data for training. During inference, a reference is optional because these models can directly use a combination of the trained tokens. But it is probably just as problematic to choose the right combination of tokens to achieve the desired target style \cite{wang2018style, valle2020mellotron} as it would be to choose the right reference utterance. Style token models are also susceptible to feature entanglement; conditioning on out-of-distribution reference utterances can lead to unstable results \cite{hsu2018hierarchical}.

Recent work has tried specifying the target prosody, emotion, or speaking style in the form of a natural language prompt. \textit{PromptTTS} trains a style encoder which takes a `style prompt' as input: a natural language description of the speaking style \cite{guo2022prompttts}. The encoder then predicts values for 5 categorical parameters (for example \textit{gender} and \textit{emotion}) to condition speech generation. The proposed method is largely inspired by the success of models like InstructGPT \cite{ouyang2022training} and allows users to specify speaking styles using only natural language. However, this method requires a training speech corpus labelled with ground-truth style prompts: very few such corpora exist. 

InstructTTS \cite{yang2023instructtts} also offers prompt-based control. The proposed method generates a latent speaking-style representation from the speech, text, and a ground-truth natural language style prompt. This representation is used to condition the text encoder and a diffusion-based \cite{ho2020denoising} decoder. Like PromptTTS, InstructTTS allows for defining novel speaking styles using natural language and also requires a prompt-labelled corpus for training. 

In \cite{Yoon2022LanguageME}, a GPT-3 model \cite{brown2020language} predicts an emotion label and intensity from text. These predicted features are then used to condition speech generation. Although the proposed method allows for predicting these features directly from text, it still requires emotion-labelled data for training.

\section{Method}
Our TTS model is based on the FastSpeech-2 \cite{renfastspeech} architecture which explicitly models \F0, energy and duration but has no reference or prompt encoder; it is further described in Section \ref{sec:model}. We propose to prompt an LLM to suggest modifications to \F0, energy and duration, based on the target text and optional contextual information, as described in Section \ref{sec:modification_method}. Predicting relative changes is assumed to be a simpler task than absolute values, which turns out to be important as discussed in Section \ref{sec:prompting}.

\subsection{Model architecture} \label{sec:model}
We use a slightly modified FastSpeech-2 \cite{renfastspeech} as our \textbf{baseline} TTS architecture. Briefly, FastSpeech-2 comprises a phoneme encoder, a variance adaptor, and a mel-spectrogram decoder. The encoder and decoder each comprise 4 feed-forward transformer blocks. In FastSpeech-2, the variance adaptor predicts \F0 and energy (after predicting each phone's duration). Energy prediction is dependent on the predicted value of \F0. We wish to modify both and we choose to do this per word, whereas FastSpeech-2 predicts per frame. We therefore replace the FastSpeech-2 variance adapters and duration model with the \textit{low-level prosody predictor} and Gaussian upsampling module from Daft-Exprt \cite{zaidi22b_interspeech}. 

\begin{figure}[t]
  \centering
\includegraphics[width=0.8\linewidth]{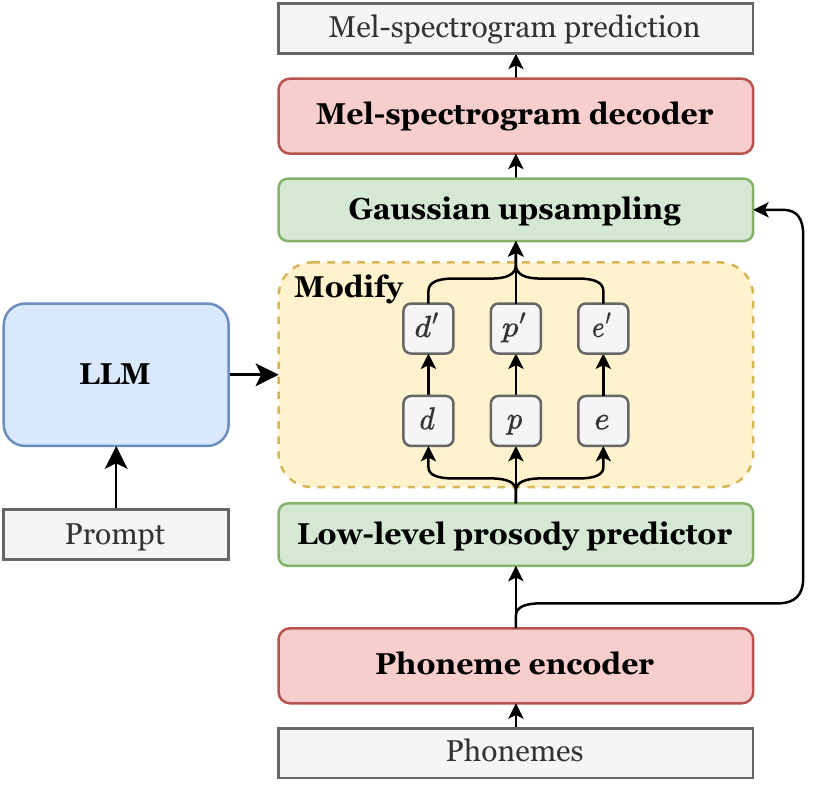}
  \caption{The proposed method adjusts the acoustic features predicted by our baseline model architecture. During inference, the low-level prosody predictor predicts per-phone duration, \F0 and energy; those initial values are then modified before continuing the forward pass.}
  \label{fig:diagram}
\end{figure}

The low-level prosody predictor predicts phone durations and per-phone speaker-normalised log-\F0 and log-energy. The baseline model architecture predicts these all together; we can then modify them independently. The Gaussian upsampling module predicts a duration distribution for each phone based on the encoder output and the previously predicted \F0, energy and duration. The module then samples a phone duration from each of those distributions. Per-phone \F0 and energy predictions are projected and summed with the encoder output, then upsampled using these phone durations. This upsampled latent representation is then used by the decoder to predict mel-spectrogram frames. During training, the model uses ground-truth values for duration, \F0, and energy. Our proposed method uses this baseline architecture but makes modifications to the predicted phone-level acoustic features as shown in Figure \ref{fig:diagram}.

\subsection{Modification method} \label{sec:modification_method}
Given the sequences of per-phone durations, \F0, and energy for an input sequence of length $T$ $(\mathbf{d}{=}d_{1:T}, \mathbf{p}{=}p_{1:_T}, \mathbf{e}{=}e_{1:T})$ predicted by the TTS model, we wish to find more appropriate values ($\mathbf{d}^{\prime}$, $\mathbf{p}^{\prime}$, $\mathbf{e}^{\prime}$). We propose a very simple 2-level modification procedure to control both utterance (`global') and word (`local') prosodic effects. More sophisticated schemes would be possible, but this suffices as a proof of concept. Global modifications may convey, for example, some aspects of emotion, attitude or speaking style. Local modifications may, for example, choose between different meanings. The method does not directly model any particular emotions or styles, nor does it require data annotated with them.

Global modifications are applied to all phones in the utterance using three coefficients. Durations and energies are scaled by $G_d$ and $ G_e$ respectively. Global \F0 scaling resulted in artefacts so instead we shift \F0 values by $G_p$. $G_d$ and $G_e$ are  limited to $[0.5, 2]$ while $G_p$ is limited such that $p^{\prime}_i$ remains within the natural \F0 range of the target speaker. Local (word) modifications are realised using three coefficient sequences, $\delta_{1:W}$, $\pi_{1:W}$, and $\epsilon_{1:W}$ for an input text consisting of $W$ words. The values resulting from both global and local modifications for phone $i$ appearing in word $j$ are:
\begin{align}
    d^{\prime}_i &= d_i \cdot G_d \cdot \delta_j, & G_d \in [0.5, 2], \quad \delta_j \in [1.0, 2.0] \label{eq:dur} \\
    e^{\prime}_i &= e_i \cdot G_e \cdot \epsilon_j, & G_e \in [0.5, 2], \quad \epsilon_j \in [1.0, 2.0] \label{eq:energy} \\
    p^{\prime}_i &= p_i + G_p + \pi_j, & G_p+\pi_j \in [p_{\text{min}}, p_{\text{max}}] \label{eq:pitch}
\end{align}
where $p_\text{min}$ and $p_\text{max}$ are the minimum and maximum allowed changes in \F0 determined by the corpus statistics. 

The low-level prosody predictor predicts log-scale and speaker-normalised energy and \F0. So, before applying the modifications in Equations \ref{eq:energy} and \ref{eq:pitch} we first convert them to linear scale and de-normalise. After applying the modifications, we re-normalise and take the log before passing them to the Gaussian upsampling module. We do not make any changes to \F0 or energy of unvoiced phones and do not modify pauses (changing pause durations resulted in considerable artefacts, possibly caused by inaccurate alignments). After the acoustic model predicts the initial values of the acoustic parameters we make the modifications determined by the chosen local and global coefficients, as shown in Figure \ref{fig:diagram}. An expert could potentially determine appropriate values for these coefficients, but here we prompt an LLM to suggest them.

\subsection{Prompt-engineering for acoustic coefficient prediction} \label{sec:prompting}
We used InstructGPT \cite{ouyang2022training} via the OpenAI API\footnote{\url{https://openai.com}}. InstructGPT is a fine-tuned version of GPT-3 \cite{brown2020language}, a 175 billion parameter autoregressive language model trained on over 400 billion tokens. InstructGPT has been fine-tuned using a mixture of supervised training and reinforcement learning, to follow instructions supplied to it through a natural language prompt \cite{ouyang2022training}. In the current work, the goal of the LLM is to generate appropriate values for the global and local coefficients given the target text. For this task, the prompt consists minimally of a description of the task and the target text. However, using minimal prompts resulted in inconsistent and nonsensical responses. Here we explain which text-based instructions we include in our prompt\footnote{A full example prompt is available at \url{https://atlisig.github.io/prompt-to-speech/index.html}} to get more consistent predictions from the model.

The ranges from which global and local coefficients are drawn, shown in Equations \ref{eq:dur}-\ref{eq:pitch}, are not the same. We found that the LLM struggled with consistently predicting values within an appropriate range for each coefficient. Because of this, we instruct the model to predict global values in the range $[-5,5]$ and local values in $[0, 5]$. We then linearly map these values to the appropriate range for each coefficient. We explain in the prompt what each coefficient controls and what a negative or positive value represents. We found that a mixture of few-shot \cite{min2022rethinking} and chain-of-thought \cite{wei2022chain} prompting also helped. So we include human-generated predictions in the prompt (few-shot) and, in addition, each intermediate step in generating the prediction is reasoned (chain-of-thought). We wrote 10 such examples which are included in the LLM prompt.

\begin{figure}[t]
  \centering  \includegraphics[width=0.7\linewidth]{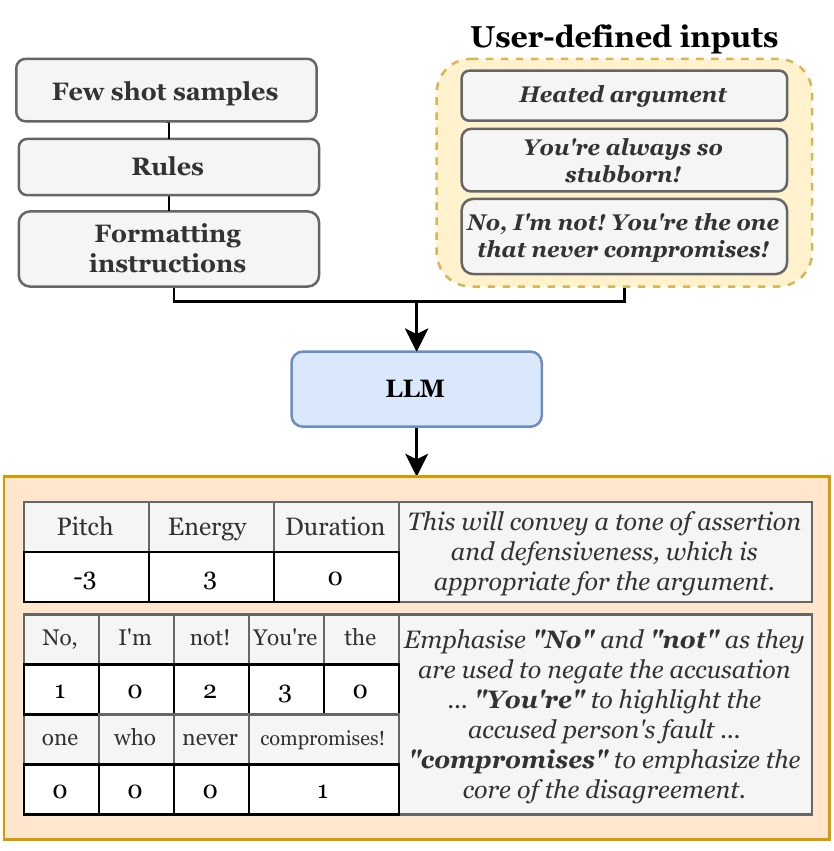}
  \caption{The LLM suggests acoustic modifications, given the target text and, optionally, a speaking style or previous dialogue context. The rules, supplied to the LLM in the prompt, ask the LLM produce reasoning for the modifications.}
  \label{fig:instruct_pipeline}
\end{figure}

We also use a set of rules that the LLM was instructed to follow. For example, the model is told to predict the parameters independently of the target voice. We also found that the LLM would often skip words from the target text in its prediction, and often make up new ones instead. But we found that giving the LLM strict formatting instructions helped with this. Together, the task description, the chain-of-thought examples, the rules, and the formatting instructions form the prompt that is supplied to the LLM \textit{only in the form of natural language}. The complete procedure is illustrated in Figure \ref{fig:instruct_pipeline}. The flexibility of the method means that we can prompt the LLM to make predictions based just on text or with additional contextual information such as speaking style or dialogue. Our prompt includes clear instructions which task exactly we want the LLM to solve.

\section{Experiments}
\subsection{Evaluation setup}
We trained all models on LJSpeech \cite{ljspeech17}. The LJSpeech corpus is single-speaker and relatively inexpressive. We remove utterances shorter than 1.5 seconds and use close to 11~000 utterances for training. 80 bin mel spectrograms are extracted from \SI{22050}{\Hz} waveforms. Phone alignments are found using the Montreal forced aligner \cite{mcauliffe2017montreal}. \F0 is estimated using REAPER\footnote{\url{https://github.com/google/REAPER}} and we use the $l^2$-norm of spectrogram frames for energy.

We compared the \textbf{proposed} method to several other models. The slightly modified FastSpeech-2 model, as described in Section \ref{sec:model}, serves as the \textbf{baseline} comparison model. Since LJSpeech is relatively inexpressive, listeners might be biased in favour of \textit{any} acoustic variance when listening to utterances whose propositional content indicates an expressive speaking style. We therefore compared the proposed method to \textbf{random}, where modification coefficients are pseudo-randomly drawn from an appropriate distribution for each coefficient. We also compared the proposed method to two reference-based oracles. We used Daft-Exprt \cite{zaidi22b_interspeech} as a representative PT model since its architecture is derived from FastSpeech-2. We also included a style-token model, in which the style token layer from \cite{wang2018style} is added to the reference encoder in Daft-Exprt. We performed inference with the style-token model using reference conditioning as described in \cite{wang2018style}. We denoted these models as oracles, \textbf{$\text{Oracle}_{\text{PT}}$ }and \textbf{$\text{Oracle}_{\text{GST}}$}, since they have full access to the ground truth target mel-spectrogram.

Each models was trained for 24 hours on 8 NVIDIA A100-SXM-80GB GPUs with a batch size of 192. We also fine-tuned a pre-trained HiFi-GAN vocoder \cite{kong2020hifi} on mel-spectrograms generated by each model we trained, thus creating a matched vocoder for each model. We use hyper-parameter configurations suggested in the original work for all models we used. 

We evaluated perceived naturalness with a Mean Opinion Score design (MOS) and appropriateness with a preference A/B/C design. We recruited native English-speaking listeners via Prolific \footnote{\url{https://www.prolific.co}} and each screen is evaluated by 8 different listeners.

\subsection{Perceived naturalness in neutral discourse}
We evaluated the perceived naturalness of the proposed method in a neutral discourse task. Here the LLM is prompted to only make local modifications to appropriately emphasise words in the text. We compared the proposed method with the \textbf{baseline} model and the two reference-based \textbf{oracles}. We evaluated 30 unseen utterances from LJSpeech, resulting in $30{\times}8{=}240$ evaluations per model. The results are shown in Table \ref{tab:mos-results}. Ground truth utterances were perceived significantly better than all models according to a paired t-test ($p{<}.05$). The proposed method is comparable to both the \textbf{baseline} ($p{=}.27$) and \textbf{$\text{Oracle}_{\text{GST}}$} ($p{=}.12$).

This task represents the optimal case for the two \textbf{oracles} since we have an appropriate reference for conditioning; namely same-text ground-truth utterances spoken by the training voice. Under these conditions, \textbf{$\text{Oracle}_{\text{PT}}$} was rated better than all other models, including the proposed method. We take the results presented in
Table \ref{tab:mos-results} as an indication that under these optimal conditions, guiding prosody with a reference utterance is indeed better than the method proposed. However, synthesising particular speaking styles or contextually appropriate prosody using reference-based models requires finding an appropriate reference for conditioning. Finding such a reference is a non-trivial task when the speech corpus is not style-labelled. Furthermore, most speech corpora are limited in the types of speaking styles that the voice performs. This eliminates the \textbf{oracles} from the two other tasks we evaluate in Sections \ref{sec:target_style}-\ref{sec:dialogue}; underlining the limitations of reference-based models. 

\begin{table}
\caption{MOS results for the neutral discourse task.} \label{tab:mos-results}
\centering
\begin{tabular}{lc}
\toprule
\multicolumn{1}{l}{\textbf{Model}} & \multicolumn{1}{c}{\textbf{Naturalness MOS}} \\
\midrule
Baseline                                & $3.1\pm0.2$             \\
Proposed                                & $3.1\pm0.2$             \\
$\text{Oracle}_{\text{GST}}$            & $3.2\pm0.2$             \\
$\text{Oracle}_{\text{PT}}$             & $3.5\pm0.2$             \\
\midrule
Ground truth                            & $4.1\pm0.1$             \\
\bottomrule
\end{tabular}
\end{table}

\subsection{Target style task} \label{sec:target_style}
Perceived \textit{appropriateness} was used to evaluate the proposed method in predicting prosody for a given target speaking style. Participants were shown the target text and speaking style when performing this evaluation. We created a small text corpus for this where we first selected a number \textit{speaking styles} and, for each one, then generated a set of target texts that we deem would be appropriate for that style. Here we define a speaking style as a distinct manner of speaking appropriate in a particular context. The flexibility of this approach means that we can define highly specific speaking styles to evaluate. Our list included styles such as \textit{``frightened''}, \textit{``in a hurry''} and \textit{``speaking to a child''}. They were chosen solely on the basis that they likely correspond with different settings of the modification coefficients.

\begin{table}
\caption{A/B/C appropriateness preference results for both the target style and dialogue tasks.} \label{tab:preference-results}
\centering
\begin{tabular}{lccc}
\toprule
\textbf{Task} & \textbf{Proposed} & \textbf{Baseline} & \textbf{Random} \\
\midrule
Target style              & 51.4\%            & 30.9\%            & 17.7\%\\
Dialogue                  & 48.4\%            & 31.0\%            & 20.6\% \\
\midrule
Mean                  & 49.9\%            & 31.0\%            & 19.1\% \\
\bottomrule
\end{tabular}
\end{table}

We compared the proposed method to \textbf{baseline} and \textbf{random} using an A/B/C preference test. We created 7 different speaking styles and 10 target texts for each one, yielding $7{\times}10{\times}8{=}560$ sets of stimuli. Participants were asked to base appropriateness on how well they thought the overall quality ofd the voice, emotion and attitude fit the target text. The results for this task are shown in the first row in Table \ref{tab:preference-results}. In the majority of cases, the proposed method is rated as most appropriate and was preferred at a ${>}70\%$ higher rate than \textbf{baseline}. Raters did not indicate a high preference for \textit{random}, suggesting that random variation is not enough to bias listeners.

\subsection{Dialogue task} \label{sec:dialogue}
\begin{figure}[t]
  \centering
  \includegraphics[width=\linewidth]{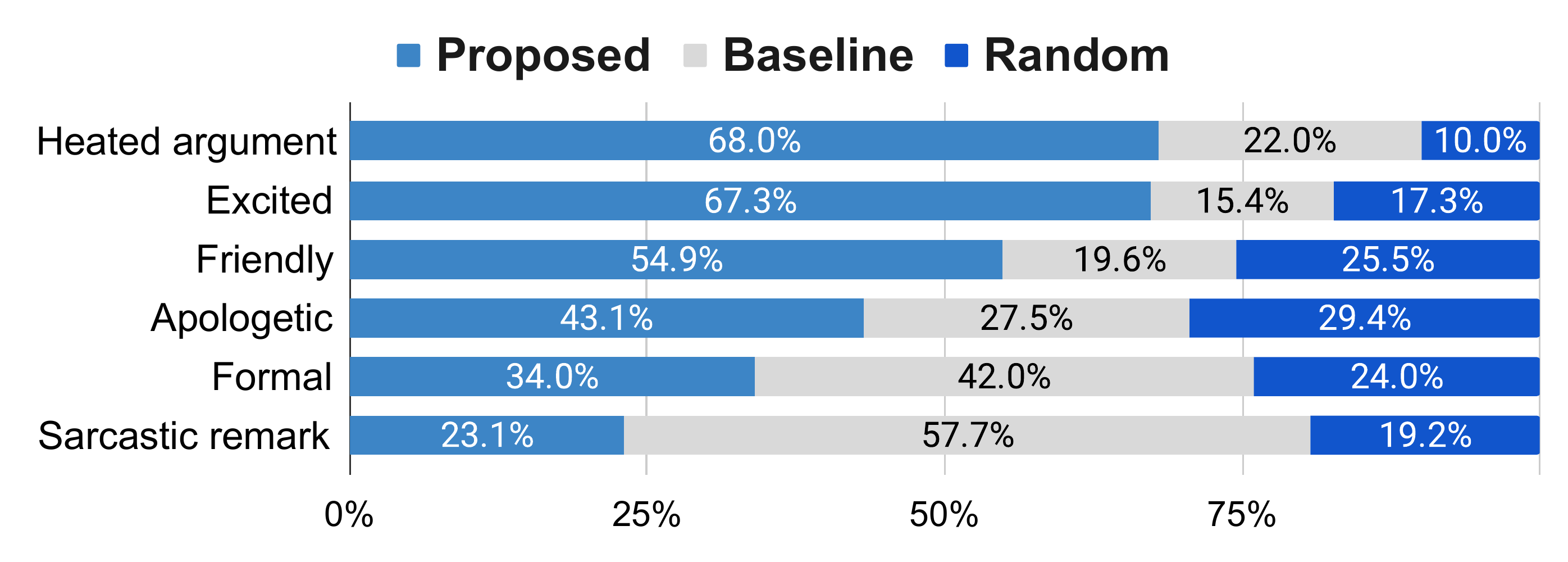}
  \caption{Preference results for the dialogue task.}
  \label{fig:dialogue_results}
\end{figure}

We also evaluated the appropriateness of utterances in the context of a two-line dialogue. Here, the LLM is instructed to make modifications based only on the previous line in a dialogue and the target text, \textit{not} a target style. We first chose six \textit{hidden} speaking styles and then generated 10 two-line dialogues where the target text would naturally fit the speaking style. The 6 styles we chose for this task are shown in Figure \ref{fig:dialogue_results} and an example dialogue is shown in Figure \ref{fig:instruct_pipeline}. We evaluated the proposed method again using an A/B/C preference test against the \textbf{baseline} and \textbf{random} with  $6{\times}10{\times}8{=}480$ sets of stimuli in total. Participants indicated a clear preference for the proposed method as shown in the second row in Table \ref{tab:preference-results}. Results in Figure \ref{fig:dialogue_results} show that the proposed method is better suited for more expressive styles, such as \textit{heated argument} or \textit{excited}, but less so for more neutral styles such as \textit{formal discussion}.

\section{Discussion}
The results demonstrate that the proposed method can adjust prosody to improve listeners' identification of novel speaking styles (i.e., not represented in the training data). The flexibility of the method allows for describing highly specific styles, or even dialogue contexts, using only natural language prompts. Future work should focus on a more sophisticated modification method (not just scaling and shifting values per utterance and per word). Future work should also consider different languages: the LJSpeech speaker is US American and the LLM used in this work was trained on 97\% English data\cite{brown2020language}, so we cannot draw any conclusions for other languages.

\clearpage
\bibliographystyle{IEEEbib}
\bibliography{refs}

\end{document}